\documentclass[sigconf]{acmart}
\usepackage{multirow}    
\usepackage{booktabs}  
\definecolor{darkred}{RGB}{139,0,0}  
\definecolor{darkblue}{RGB}{0,0,139}  

\setcopyright{acmlicensed}
\copyrightyear{2025}
\acmYear{2025}
\setcopyright{acmlicensed}
\acmConference[MM '25] {Proceedings of the 33rd ACM International Conference on Multimedia}{October 27--31, 2025}{Dublin, Ireland.}

\begin{document}

\title[LSFDNet]{LSFDNet: A Single-Stage Fusion and Detection Network for Ships Using SWIR and LWIR}


%
%

\author{Yanyin Guo}
\orcid{0009-0000-0515-0321}
\affiliation{%
  \institution{Zhejiang University}
  \city{Hangzhou}
  \state{Zhejiang}
  \country{China}
}
\email{guoyanyin@zju.edu.cn}

\author{Runxuan An}
\orcid{0009-0001-8674-5008}
\affiliation{%
  \institution{Zhejiang University}
  \city{Hangzhou}
  \state{Zhejiang}
  \country{China}
}
\email{22331002@zju.edu.cn}

\author{Junwei Li}
\authornote{Corresponding Author.}
\orcid{0000-0001-6957-3059}
\affiliation{%
 \institution{Zhejiang University}
 \city{Hangzhou}
 \state{Zhejiang}
 \country{China}
}
\email{lijunwei7788@zju.edu.cn}

\author{Zhiyuan Zhang}
\orcid{0000-0003-3945-5638}
\affiliation{%
  \institution{Singapore Management University}
  \city{Singapore}
  \country{Singapore}
}
\email{cszyzhang@gmail.com}

\renewcommand{\shortauthors}{Yanyin Guo, Runxuan An, Junwei Li, Zhiyuan Zhang}

\begin{abstract}
Traditional ship detection methods primarily rely on single-modal approaches, such as visible or infrared images, which limit their application in complex scenarios involving varying lighting conditions and heavy fog. To address this issue, we explore the advantages of short-wave infrared (SWIR) and long-wave infrared (LWIR) in ship detection and propose a novel single-stage image fusion detection algorithm called LSFDNet. This algorithm leverages feature interaction between the image fusion and object detection subtask networks, achieving remarkable detection performance and generating visually impressive fused images. To further improve the saliency of objects in the fused images and improve the performance of the downstream detection task, we introduce the Multi-Level Cross-Fusion (MLCF) module. This module combines object-sensitive fused features from the detection task and aggregates features across multiple modalities, scales, and tasks to obtain more semantically rich fused features. Moreover, we utilize the position prior from the detection task in the Object Enhancement (OE) loss function, further increasing the retention of object semantics in the fused images. The detection task also utilizes preliminary fused features from the fusion task to complement SWIR and LWIR features, thereby enhancing detection performance. Additionally, we have established a Nearshore Ship Long-Short Wave Registration (NSLSR) dataset to train effective SWIR and LWIR image fusion and detection networks, bridging a gap in this field. We validated the superiority of our proposed single-stage fusion detection algorithm on two datasets. The source code and dataset are available at \href{https://github.com/Yanyin-Guo/LSFDNet} {https://github.com/Yanyin-Guo/LSFDNet}. 
\end{abstract}




\maketitle
\begin{figure}[htbp]
    \centering
    \includegraphics[width=\linewidth]{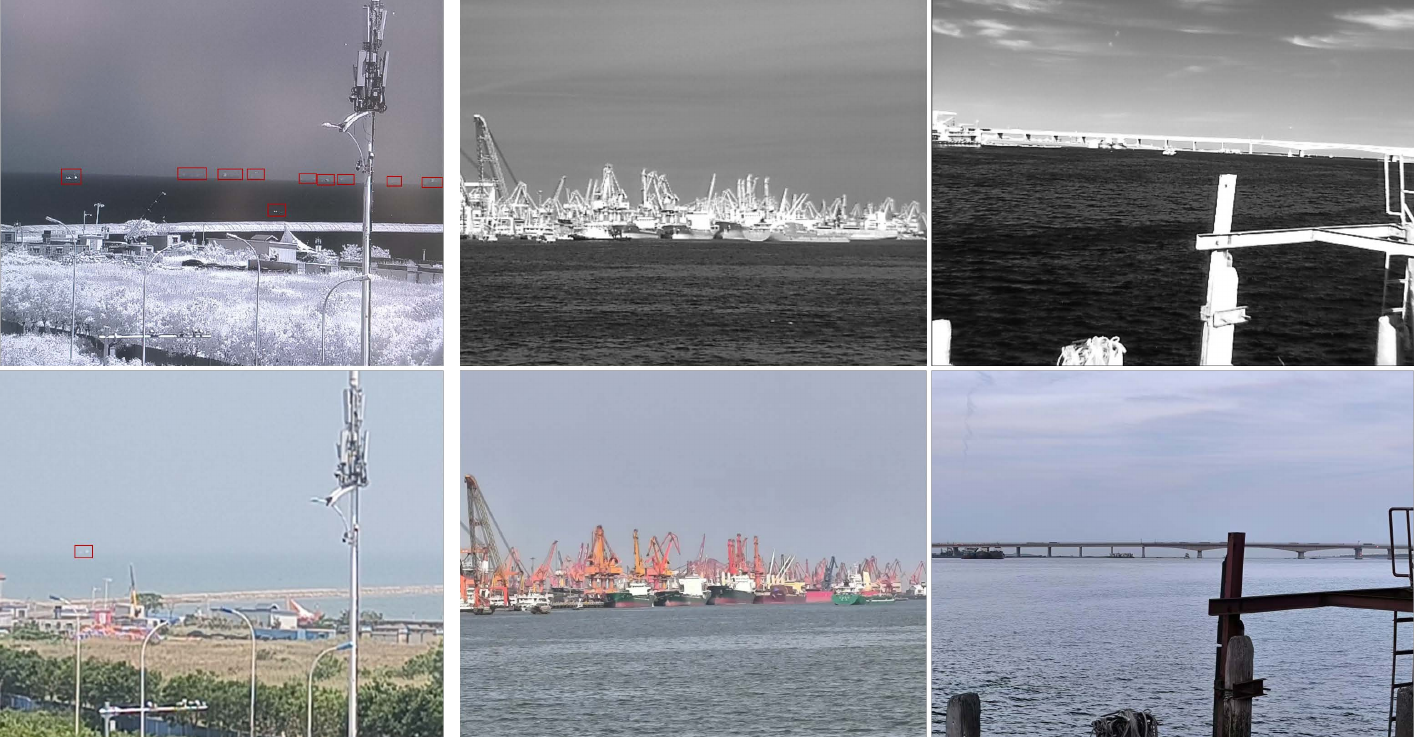}
    \begin{flushleft}  {\fontsize{8pt}{10pt}\selectfont  
    \vspace{-3pt}
    \hspace{0.2cm} (a) fog penetration \hspace{1.60cm} (b) sea surface contrast} \end{flushleft}
    \vspace{-5pt}
    \caption{SWIR (Top) vs. Visible (Bottom) Imaging.}\label{fig:SWIR}
    \Description{comparison}
\end{figure}
\section{Introduction}
\label{sec:intro}
\begin{figure}[t]
    \includegraphics[width=\linewidth]{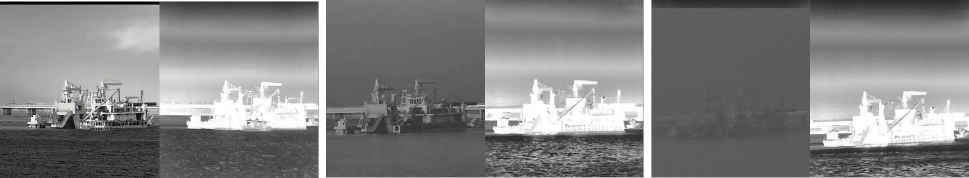}
    \begin{flushleft}  {\fontsize{8pt}{10pt}\selectfont  
    \vspace{-3pt}
    \hspace{0.43cm} Normal lighting \hspace{1.30cm} Evening \hspace{1.55cm} Nighttime \\ } \end{flushleft}
    \vspace{-5pt}
    \caption{Comparison of infrared radiation signals between SWIR (Left) and LWIR (Right) images.}\label{fig:light}
    \Description{light}
\end{figure}
Ship detection plays a pivotal role in modern maritime technology, with applications spanning maritime safety, port management, and other related fields~\cite{khan2023deep, rekavandi2025guide, deng2025trend}. Most existing approaches rely heavily on single-modal object detection, such as visible or infrared images~\cite{wu2023ship, rekavandi2025guide, wang2020ship}. However, these methods frequently struggle with reduced accuracy in challenging maritime conditions, including heavy fog, sea surface waves, and variable lighting.

Compared to visible light, short-wave infrared (SWIR) imaging (0.9–2.5 $\mu$m) leverages the reflection of SWIR radiation from objects, delivering unique benefits for maritime target detection~\cite{wang2024lightweight, xiangyue2022ship}: (1) Superior signal-to-noise ratio (SNR) under adverse conditions. As shown in Fig.~\ref{fig:SWIR} (a), SWIR can penetrate thin fog, smoke, and other aerosol particles, maintaining high image clarity even in low-visibility environments, which makes it well suited for challenging maritime conditions. (2) High contrast against the sea surface. Fig.~\ref{fig:SWIR} (b) illustrates that seawater almost completely absorbs SWIR radiation, resulting in a dark background, while weak targets reflecting SWIR appear significantly brighter. This high contrast facilitates easier object detection. (3) Efficient image processing. SWIR images are single-channel and similar to grayscale images in the visible spectrum, simplifying image processing workflows.

These advantages make SWIR imaging highly effective for maritime object detection and recognition~\cite{bao2018automatic,firdiantika2024yolo}. Nonetheless, SWIR imaging has its limitations, as most daytime SWIR radiation originates from sunlight. As shown in Fig.~\ref{fig:light}, the intensity of ships in SWIR images depends strongly on solar irradiance. Thus, under low-light conditions like overcast days or dusk, ship brightness diminishes and target-water contrast drops. Conversely, long-wave infrared (LWIR, 8–14 $\mu$m) imaging relies on thermal radiation emitted by objects, rendering it less susceptible to lighting variability. However, LWIR images typically lack fine details and texture, as depicted in Fig.~\ref{fig:light}.

It is evident that single-band infrared imaging is susceptible to environmental interference. SWIR images retain texture details but are sensitive to illumination changes, while LWIR images are robust to lighting but tend to lose edge details owing to thermal diffusion and surface noise. To fully exploit the complementary features of these two modalities, a fused image can simultaneously retain thermal radiation information (from LWIR) and texture details (from SWIR), making it more suitable for ship detection in complex scenarios. However, existing works mainly focus on visible-infrared image fusion~\cite{yuan2022translation, tang2024drmf, li2023learning}, with limited research devoted to fusing LWIR and SWIR modalities for maritime applications. And practical datasets are also scarce. Furthermore, traditional cascade networks treat fusion and detection as decoupled, two-stage tasks, often prioritizing fused image quality while overlooking task-specific improvements for downstream detection. This separation limits the interaction between the fusion and detection processes, hindering the effective use of complementary information from both modalities.

To address these challenges, we propose LSFDNet, a single-stage Long-Short Wave Fusion Detection Network for robust ship detection. The core of LSFDNet is a Multi-Level Cross-modal Fusion (MLCF) module to exploit complementary features of LWIR and SWIR images across three dimensions: (1) cross-modal complementarity, integrating effective features from SWIR and LWIR images; (2) multi-scale complementarity, leveraging hierarchical feature representations at different granularities; and (3) task complementarity, enabling semantic interaction between fusion and detection tasks. These are achieved via three Multi-Feature Attention blocks, which combine self-attention and cross-modal attention to aggregate pixel-level information across modalities. To further enhance scene understanding and highlight ship targets, LSFDNet incorporates task-specific semantic features from the detection branch via residual connections. For the fusion task, we introduce object-location priors and a novel Object-Enhancement (OE) loss function, which encourages the fused images to emphasize target-relevant semantics. In parallel, for the detection task, fused modality features are used to augment unimodal representations, leading to improved detection accuracy. LSFDNet is trained in an end-to-end manner, where jointly optimized loss functions guide both fusion and detection branches, ensuring high-quality fusion while maximizing detection performance. This unified framework enables LSFDNet to achieve robust and accurate ship detection in complex maritime environments. Furthermore, to support research in this domain, we release the Nearshore Ship Long-Short Wave Registered (NSLSR) Dataset, a practical and aligned dataset for LWIR-SWIR fusion, fostering further exploration in multimodal maritime perception.

The main contributions of our work are summarized as follows:
\begin{itemize}
\item[$\bullet$] \textbf{Pioneering Approach}: We are the first to explore how the fusion of SWIR and LWIR images enhances ship detection, emphasizing the unique role and potential of SWIR in maritime detection tasks.
\item[$\bullet$] \textbf{Integrated Fusion-Detection Architecture}: We propose LSFDNet, a single-stage network that seamlessly combines image fusion and object detection into a unified, end-to-end framework. Cross-task feature interactions improve both visual fidelity and detection performance.
\item[$\bullet$] \textbf{Advanced Feature Aggregation and Task-Specific Loss Function}: We design the Multi-Level Cross-Fusion (MLCF) module to effectively aggregates features across modalities, scales, and tasks. Additionally, we design an Object Enhancement (OE) loss, which leverages positional priors from detection to enhance the visibility of ships in fused images while suppressing sea surface noise. 
\item[$\bullet$] \textbf{New Dataset}: We introduce the Nearshore Ship Long-Short Wave Registration (NSLSR) dataset, which consists of 1,205 pairs of well-registered SWIR and LWIR images with 2,818 annotated objects. The dataset captures a variety of complex coastal scenarios with diverse lighting conditions, making it a valuable resource for maritime detection research.
\end{itemize}

\section{Related Works}
\label{sec:related}
\subsection{Infrared Ship Detection} \label{Infrared Ship Detection}
Infrared imaging technology has demonstrated remarkable advantages in ship detection due to its ability to maintain stable imaging quality even in complex maritime environments and low-light conditions. In recent years, research on ship detection algorithms based on infrared images has made significant progress~\cite{firdiantika2024yolo, guo2024yolo, wu2023mtu, wang2020ship}. In 2024, Wang et al.~\cite{li2022ggt} introduced an innovative attention-based feature fusion module and the SPD-Conv algorithm, significantly enhancing the model's performance in detecting small objects and densely arranged ships. In the same year, Guo et al.~\cite{guo2024multi} proposed the MAPC-Net model, which incorporates multi-scale attention mechanisms within a multi-scale feature pyramid network to further optimize detection performance. Subsequently, the same team introduced the FCNet model~\cite{guo2024fcnet}, combining the strengths of dilated convolutions and deformable convolutions, achieving a new breakthrough in detection accuracy and performance. In 2025, Wang et al.~\cite{wang2025ppgs} developed the lightweight PPGS-YOLO network to cater to the specific requirements of nearshore application scenarios, while Zhao et al.~\cite{zhang2021infrared} addressed the challenge of missing details in infrared images by leveraging attention mechanisms and local convolutional interactions to effectively enhance the feature saliency of weak objects.
\subsection{Image Fusion and Object Detection} \label{Image Fusion and Object Detection}
Multimodal image fusion integrates complementary information from different sensors to enhance both the visual quality and the semantic richness of images. Recently, learning-based fusion methods have made significant advancements, with innovations primarily focusing on the optimization and design of network architecture~\cite{lu2024dcafuse, tang2024itfuse, li2024crossfuse, xu2025hifusion} as well as improved feature representation strategies~\cite{huang2024efficient, yang2024cefusion, zhang2025illumination}. In the latest studies, Xiao et al.~\cite{xiao2024fafusion} introduced a frequency-aware learning mechanism, proposing a frequency-aware network tailored for infrared and visible image fusion. Zheng et al.~\cite{zheng2024frequency} developed a novel architecture combining frequency integration and spatial compensation. Tang et al.~\cite{tang2024drmf} applied diffusion models to the image fusion, effectively mitigating information degradation during the fusion process. In terms of network design, Wang et al.~\cite{wang2024terf} constructed a region-aware fusion framework supporting interactions between text and visual models. Liu et al.~\cite{liu2025three} proposed a fusion strategy based on three-dimensional features, expanding the feature space by capturing common characteristics of scenes.

Recent research has begun to jointly optimize image fusion with downstream tasks, such as segmentation~\cite{tang2022superfusion, tang2023rethinking} and fusion with detection. For fusion-detection tasks, Liu et al.~\cite{liu2022target} proposed a dual-level optimization framework and designed a target-aware dual-path adversarial learning network to optimize both fusion and detection. Similarly, Sun et al.~\cite{sun2022detfusion} introduced a detection-driven infrared and visible image fusion network, employing a cascaded structure and using detection loss to guide the training of the fusion network via backpropagation. However, such cascaded frameworks often face high training complexity. To address this limitation, Zhang et al.~\cite{zhang2024e2e} proposed an end-to-end multimodal synchronous fusion-detection framework, which not only simplified the training process but also produced high-quality fused images and highly accurate detection results.

Notably, current research almost exclusively focuses on infrared and visible image fusion, while the fusion of LWIR and SWIR images remains an unexplored and critical area of study.

\begin{figure*}[tbp]
    \centering
    \includegraphics[width=0.95\linewidth]{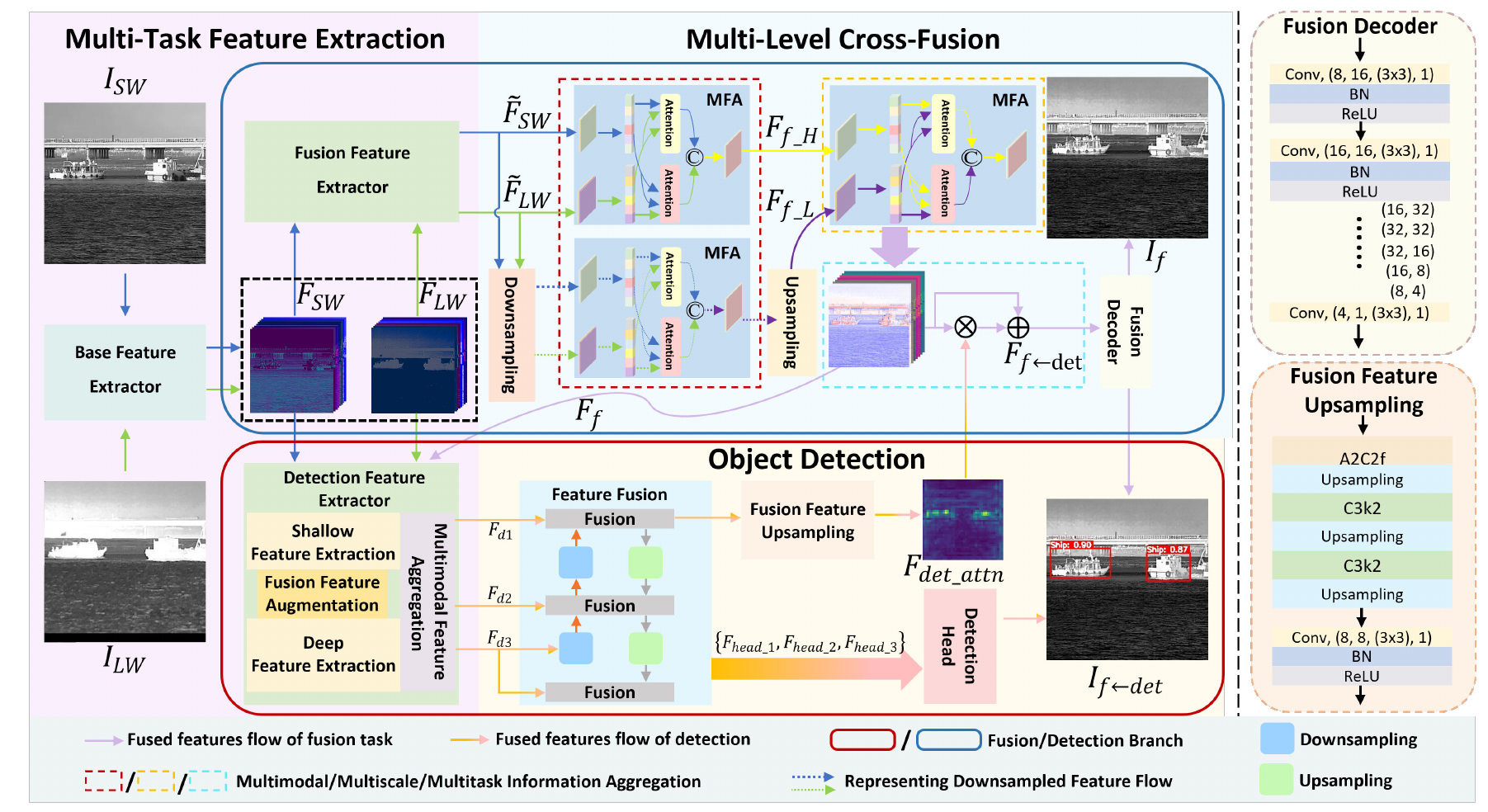}
    \vspace{-5pt}
    \caption{Overview of the proposed LSFDNet. The LSFDNet comprises a deeply coupled fusion network and a detection network. Initially, preliminary fused features $F_{f}$ are generated through multimodal and multiscale features aggregation. During the feature extraction phase, the detection branch receives the $F_{f}$ and utilizes it to supplement shallow SWIR feature $F_{SW}$ and LWIR feature $F_{LW}$ by Fusion Feature Augmentation module. The fusion branch receives the highly aggregated detection feature $F_{det\_attn}$, enabling the fusion network to focus more on the objects. Finally, the fused image $I_{f}$ reconstructed by the fusion decoder is combined with the object location information decoded by the detection head to generate the fused detection image.}\label{fig:model}
    \Description{Model Architecture}
\end{figure*}
\section{Method}
\label{sec:method}
\subsection{Overview} \label{sec:network}
The architecture of the proposed LSFDNet is shown in Fig. \ref{fig:model}, which comprises three main components: the Multi-Task Feature Extraction (MTFE) module, the Multi-Level Cross-Fusion (MLCF) branch, and the Object Detection branch. Given a pair of registered SWIR $I_{SW}$ and LWIR $I_{LW}$ images as input, the MTFE module first extracts multiple features from both modalities, which are then passed into the fusion and detection branches for cross-task interactions and aggregations, resulting in two task-specific fused features: $F_{f\leftarrow det}$ for fusion and $\left\{F_{head\_1}, F_{head\_2}, F_{head\_3} \right\}$ for detection. Notably, in the MLCF branch, the intermediate aggregated feature $F_{f}$  is fed back into the MTFE module to supplement the SWIR and LWIR detection features. Likewise, the aggregated detection feature $F_{det\_attn}$ from the detection branch is fed into the fusion branch, forcing the network to focus more effectively on the objects. Finally, the fusion decoder generates a high-quality fused image $I_{f\leftarrow det}$, while the detection head produces highly precise object bounding boxes, achieving both superior visual quality and accurate detection performance. The details of the MTFE, MLCF, and the loss function are described below.

\subsection{Multi-Task Feature Extraction} \label{sec:MTFE}
The Multi-Task Feature Extraction (MTFE) module consists of three feature extractors: the Base Feature Extractor, the Fusion Feature Extractor, and the Detection Feature Extractor. 

The input pair of SWIR ($I_{SW}$) and LWIR ($I_{LW}$) images is first passed through the Base Feature Extractor, which consists of three convolutional layers, each with a kernel size of 3×3 and a stride of 1. This extractor generates 8-channel shallow features from both SWIR and LWIR images, denoted as $F_{SW}$ and $F_{LW}$, respectively. These base features are shared across the subsequent fusion and detection tasks. 


The Fusion Feature Extractor expands the 8-channel input features to 32 channels before reducing them back to 8 channels, producing the output features $\tilde{F}_{SW}$ and $\tilde{F}_{LW}$. This design captures finer-grained features and improves feature representation. Notably, by maintaining consistent spatial dimensions during fusion, the feature sizes remain aligned with the original input, facilitating pixel-level multimodal image fusion \cite{MulFS-CAP}.

\begin{figure*}[t]
    \includegraphics[width=0.7\linewidth]{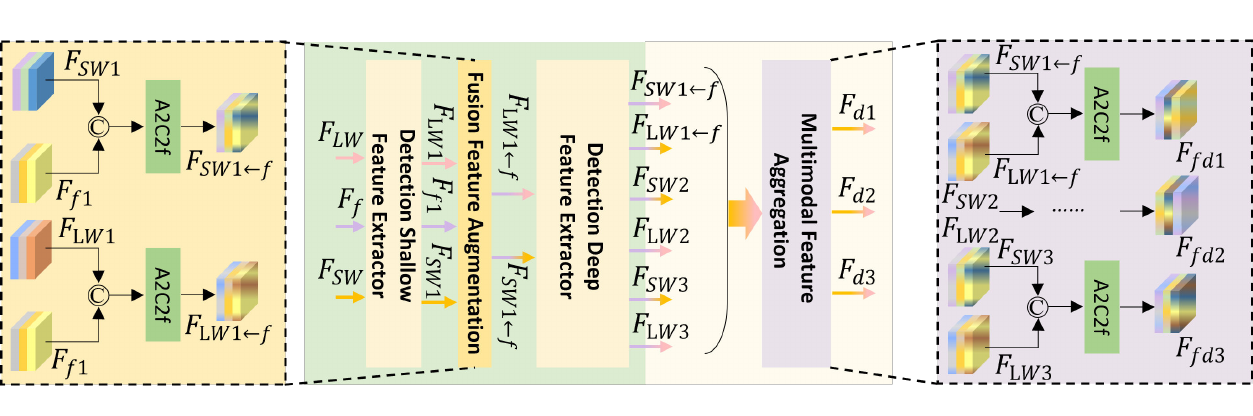}
    \vspace{-5pt}
    \caption{The network details of our Fusion Feature Augmentation and Multimodal Feature Aggregation blocks.}\label{fig:FE}
    \Description{Model Architecture}
\end{figure*}

The Detection Feature Extractor consists of four sub-modules: Shallow Feature Extraction, Fusion Feature Augmentation, Deep Feature Extraction, and Multimodal Feature Aggregation. Both Shallow and Deep Feature Extractions utilize the YOLO architecture backbone \cite{tan2025yolov12}. Shallow Feature Extraction extracts features from $F_{SW}$, $F_{LW}$, and the intermediate aggregated feature $F_{f}$ from the fusion sub-network. The Fusion Feature Augmentation module further augments $F_{SW}$ and $F_{LW}$ using $F_{f}$ to enrich their feature representations. The Multimodal Feature Aggregation module aggregates SWIR and LWIR features at multiple scales, further enhancing detection-related features. The implementation details of Fusion Feature Augmentation and Multimodal Feature Aggregation are shown in Fig. \ref{fig:FE}. Additionally, the A2C2f block \cite{tan2025yolov12} incorporates Area Attention, enabling the selection of critical features from multimodal feature maps.

\subsection{Multi-Level Cross-Fusion Module} \label{sec:MLCF}
After separately extracting features, fusion is required to combine $\tilde{F}_{SW}$ and $\tilde{F}_{LW}$ from different modalities. To ensure that the fused features comprehensively represent the scene, we design the Multi-Level Cross-Fusion (MLCF) module, which is composed of three Multi-Feature Attention (MFA) blocks. The first MFA aggregates $\tilde{F}_{SW}$ and $\tilde{F}_{LW}$ to generate cross-modal fusion features $F_{f\_H}$. At the same time, $\tilde{F}_{SW}$ and $\tilde{F}_{LW}$ are downsampled to produce lower scale multimodal features. These features are then passed through another MFA block for feature aggregation and further refined using the upsampling module to generate $F_{f\_L}$. The third MFA block combines the multi-scale feature of $F_{f\_H}$ and $F_{f\_L}$ to produce preliminary fusion features $F_{f}$. 

\begin{figure}[t]
    \centering
    \includegraphics[width=0.95\linewidth]{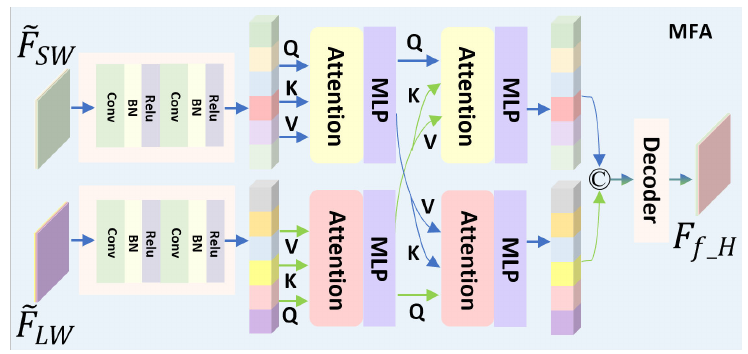}
    \caption{The network details of our Multi-Feature Attention (MFA) block.}\label{fig:MFAB}
    \Description{Model Architecture}
\end{figure}

The implementation details of the MFA block are illustrated in Fig. \ref{fig:MFAB}. First, two convolutional blocks are employed to extract and aggregate features from $\tilde{F}_{SW}$ and $\tilde{F}_{LW}$, respectively. Since image fusion operates at the pixel level, feature interactions within local regions are crucial. Therefore, the extracted features are divided into $p \times p$ small patches, which are vectorized to form $\tilde{F}_{SW}^{p}$. A linear projection is then applied to $\tilde{F}_{SW}^{p}$ to generate the corresponding $\tilde{Q}_{SW}^{p}$, $\tilde{K}_{SW}^{p}$, and $\tilde{V}_{SW}^{p}$. These are subsequently processed by a simple self-attention layer and a multilayer perceptron (MLP) to enhance the information representation within the sequence, producing the output $\overline{F}_{SW}^{p}$: \begin{equation}
  \overline {F}_{SW}^{p} = \tilde{F}_{SW}^{p} + softmax\left( \frac{\tilde{Q}_{SW}^{p} \left( \tilde{K}_{SW}^{p} \right)^{T}}{\sqrt{d_{p}}} \right) \tilde{V}_{SW}^{p},
\end{equation}where $\sqrt{d_{p}}$ represents the dimension of $\tilde{K}_{SW}^{p}$. Similarly, we can obtain $\overline {F}_{LW}^{p}$ as follows: \begin{equation}
  \overline {F}_{LW}^{p} = \tilde{F}_{LW}^{p} + softmax\left( \frac{\tilde{Q}_{LW}^{p} \left( \tilde{K}_{LW}^{p} \right)^{T}}{\sqrt{d_{p}}} \right) \tilde{V}_{LW}^{p}.
\end{equation}

To enhance feature interactions between $\overline{F}_{SW}^{p}$ and $\overline{F}_{LW}^{p}$, we introduce a cross-attention mechanism and an MLP layer. $\overline{Q}_{SW}^{p}$ is computed from $\overline{F}_{SW}^{p}$, while the $\overline{K}_{LW}^{p}$ and $\overline{V}_{LW}^{p}$ are derived from $\overline {F}_{LW}^{p}$. These are used as the $Q$, $K$, and $V$ inputs for the attention layer, resulting in $\overline{\overline{F}}_{SW}^{p}$: \begin{equation}
  \overline{\overline{F}}_{SW}^{p} = softmax\left( \frac{\overline{Q}_{SW}^{p} \left( \overline{K}_{LW}^{p} \right)^{T}}{\sqrt{d_{p}}} \right) \overline{V}_{LW}^{p}.
\end{equation}Similarly, we can obtain $\overline{\overline{F}}_{LW}^{p}$: \begin{equation}
  \overline{\overline{F}}_{LW}^{p} = softmax\left( \frac{\overline{Q}_{LW}^{p} \left( \overline{K}_{SW}^{p} \right)^{T}}{\sqrt{d_{p}}} \right) \overline{V}_{SW}^{p}.
\end{equation}

Then, we fold and concatenate $\overline{\overline{F}}_{SW}^{p}$ and $\overline{\overline{F}}_{LW}^{p}$, and feed the result into the Decoder block. The Decoder block consists of four convolutional layers, which first expand the features from 8 channels to 16 channels and then reduce them back to 8 channels, ultimately producing the fused feature $F_{f}$.Furthermore, $F_{f}$ is connected residually with $F_{det\_attn}$, which is processed through an attention mechanism from the detector network, ensuring that the fusion features focus more effectively on the objects. The MLCF module enriches the fused image features across multiple levels, including multimodal, multiscale, and multitask representations.

\subsection{Loss Function} \label{sec:Loss}
The overall loss function consists of fusion loss and detection loss, which can be expressed as:
\begin{equation}
  L = (1-\lambda) L_{f} + \lambda L_{det},
\end{equation}where $L_{f}$ represents the fusion loss, $L_{det}$ denotes the detection loss, and $\lambda$ is a balancing factor for the two loss terms. Specifically, $L_{det}$ employs the detection loss function from YOLO.

Traditionally, image fusion tasks aim to produce fused images that retain as much texture and luminance information as possible, maximizing the information entropy of the fused image. However, in the context of maritime ship detection, elements such as sea surface undulations, specular highlights, and glare introduce noise that degrades image quality. Additionally, it has been observed that in LWIR images, the contrast between ships and the sea background is significant. To address this, we apply gamma correction to enhance the contrast of objects in LWIR images while suppressing sea surface noise. The enhanced LWIR image, denoted as $I'_{LW}$, is defined as: \begin{equation}
  I'_{LW} = 255 \times \left(\frac{I_{LW}}{255}\right)^{\gamma},
\end{equation}where $\gamma$ is used to adjust the brightness and contrast of images.

In this study, the fusion task is designed to support the detection task by comprehensively capturing ship-related information. To enhance ship detection performance, the fused images should ideally preserve as much texture and brightness information as possible. Thus, our network fully leverages object location information from the detection task and designs an Object Enhancement (OE) loss function. This approach makes the contrast between object and the background more pronounced. The OE Loss consists of a global loss and an object loss, expressed as: \begin{equation}
  L_{f} = \left(1 - \sigma \right) L_{f}^{global} + \sigma L_{f}^{object},
\end{equation}where $L_{f}^{global}$ represents the global loss, $L_{f}^{object}$ denotes the local objective loss, and $\sigma$ is used to balance the loss terms.

$L_{f}^{global}$ uses gradient loss and intensity loss to learn the texture details and content of the source image, which can be formulated as follows:\begin{equation}
 L_{f}^{global} = \left(1 - \alpha \right) L_{global}^{grad} + \alpha L_{global}^{intensity},
\end{equation}\begin{equation}
 L_{global}^{grad} = \frac{1}{HW} \lVert \nabla I_{f} - max \left( \nabla I_{SW}, \nabla I_{th} \right)  \rVert _{1},
\end{equation}\begin{equation}
 L_{global}^{intensity} = \frac{1}{HW} \lVert I_{f} - max \left( I_{SW}, I_{th} \right)  \rVert _{1},
\end{equation}\begin{equation}
 L_{th} = mean \left( I_{SW}, I'_{LW} \right),
\end{equation}where $\nabla$ denotes the Sobel operator, and $\alpha$ is used to balance the loss terms. $L_{f}^{object}$ is similar to $L_{f}^{global}$ in that it calculates the gradient and intensity loss of all ship labels, represented as:\begin{equation}
 L_{f}^{object} = \left(1 - \beta \right) L_{object}^{grad} + \beta L_{object}^{intensity},
\end{equation}\begin{equation}
 L_{object}^{grad} = \frac{1}{n} \sum_{i=1}^{n} \frac{1}{H_{i}W_{i}} \lVert \nabla I_{f}^{i} - max \left( \nabla I_{SW}^{i}, \nabla I_{LW}^{i} \right)  \rVert _{1},
\end{equation}\begin{equation}
 L_{object}^{intensity} = \frac{1}{n} \sum_{i=1}^{n} \frac{1}{H_{i}W_{i}} \lVert I_{f}^{i} - max \left( I_{SW}^{i}, I_{LW}^{i} \right)  \rVert _{1},
\end{equation}where $n$ is the number of objects, and $i$ denotes the i-th object. Since the object does not involve the sea surface, the original image $I_{LW}$ is used.
\subsection{NSLSR Dataset} \label{sec:dataset}
Currently, there are very few datasets of registered SWIR and LWIR maritime ships. To the best of our knowledge, the only publicly available dataset for research is the Infrared Ship Dataset (ISD) \cite{CFOResearch2020} released by Shandong University. This dataset focuses on long-range (10-12 km) maritime ship detection and contains 1,044 image pairs at a resolution of  300×300. However, it includes only 28 unique ship instances and has limited scene diversity, with monotonous backgrounds. These limitations make it challenging to effectively train multimodal fusion or detection networks.

To address this gap, we construct a binocular synchronous system using both a SWIR and a LWIR camera, each with a resolution of 640×512. The SWIR camera is equipped with an uncooled InGaAs infrared focal plane array (FPA) detector with a pixel pitch of 15 $\mu \text{m}$  and a spectral response range of 0.9 to 1.7 $\mu \text{m}$. The LWIR camera uses an uncooled VOx infrared FPA detector with a pixel pitch of 12 $\mu \text{m}$  and a spectral response range of 8 to 14 $\mu \text{m}$.

We collect a substantial number of ship images from nearshore marine environments over different time periods. The rigid transformations between the SWIR and LWIR image pairs are manually corrected, and soft deformations are further aligned using a heterogeneous image registration algorithm \cite{ren2025minima}. After discarding poorly registered image pairs, we obtain a total of 1,205 registered LWIR-SWIR ship image pairs, which form the Nearshore Ship Long-Short Wave Registration Dataset (NSLSR). Fig. \ref{fig:light} provides a comparison of SWIR and LWIR ship images from our dataset, captured at different times. Additionally, all ship objects in the images are annotated, and the dataset is split into training and testing subsets with a 9:1 ratio. To the best of our knowledge, this is the first practical LWIR-SWIR dataset for maritime ship image fusion and detection.

\section{Experiment}
\label{sec:experiment}
\subsection{Dataset and Implementation Details} \label{sec:setup}
We conduct experiments on two datasets, NSLSR and ISD, both of which are used to evaluate the performance of image fusion. Due to the limited number of ship instances and monotonous backgrounds in the ISD dataset, we use only the NSLSR dataset to assess the performance of ship detection. Specifically, we construct a training set with $844$ images and a testing set with $361$ images from the NSLSR dataset. Among the testing set, $118$ images are used to evaluate the performance of the fusion network, while the entire testing set is utilized to evaluate detection performance. For the ISD dataset, we use $940$ images to build the fusion training set and $105$ images to test the fusion performance. For the fusion task, we use entropy ($EN$), spatial frequency ($SF$), standard deviation ($SD$),  sum of correlation differences ($SCD$), visual information fidelity ($VIF$) and edge-based metric ($Q_{abf}$) to evaluate fusion performance. Specifically, $EN$ and $SCD$ are used to represent the richness of information in the fusion image, while SF and $Qabf$ reflect the gradient and detail information. $VIF$ and $SD$ are used to assess the image visual quality as perceived by the human eye. Together, these six metrics provide a comprehensive evaluation of the fused image quality. The parameters $\sigma$, $\alpha$ and $\beta$ are set to $0.2$, $0.5$ and $0.5$, respectively. LSFDNet is optimized using the Adam optimizer with a learning rate of $1e^{-4}$ and linear decay to update the network parameters. Training is conducted on NVIDIA GeForce RTX $4090$ GPUs, adopting a warm-up strategy for the first $500$ iterations and running a total of $30,000$ iterations with a batch size of $8$.
\begin{figure}[htb]
    \includegraphics[width=0.47\textwidth]{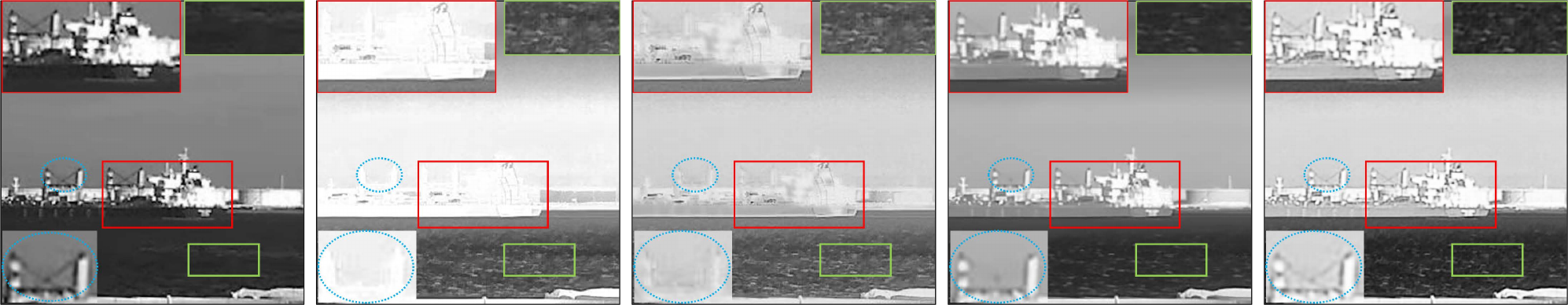}\\
    \begin{flushleft}  {\fontsize{8pt}{10pt}\selectfont  
    \vspace{-5pt}
    \hspace{0.43cm} SWIR \hspace{0.9cm} LWIR \hspace{0.75cm} DATFuse \hspace{0.58cm} DDFM \hspace{0.75cm} EMMA \\} \end{flushleft}
    \vspace{-1pt}
    \includegraphics[width=0.47\textwidth]{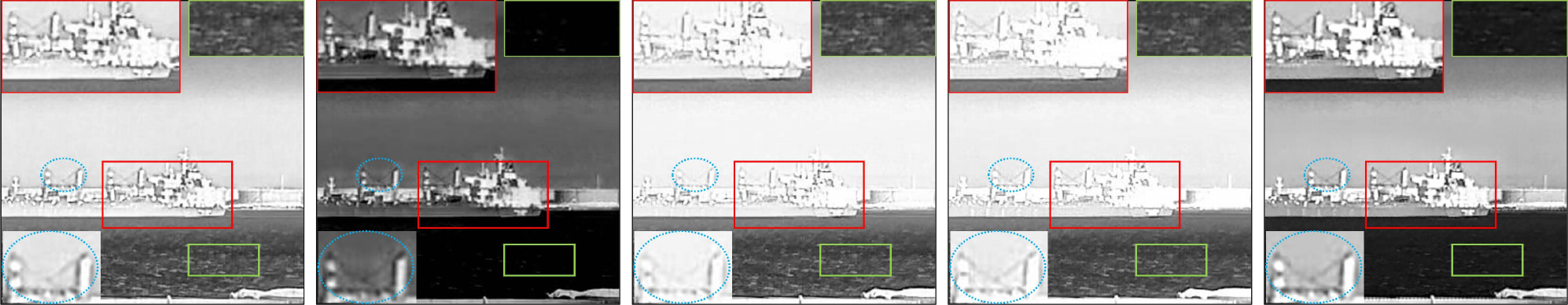}\\
    \begin{flushleft}  {\fontsize{8pt}{10pt}\selectfont  
    \vspace{-5pt}
    \hspace{0.25cm} DifFusion \hspace{0.65cm} IGNet \hspace{0.55cm} SeAFusion \hspace{0.3cm} SwinFusion \hspace{0.55cm} ours \\} \end{flushleft} 
    \vspace{-1pt}
    \includegraphics[width=0.47\textwidth]{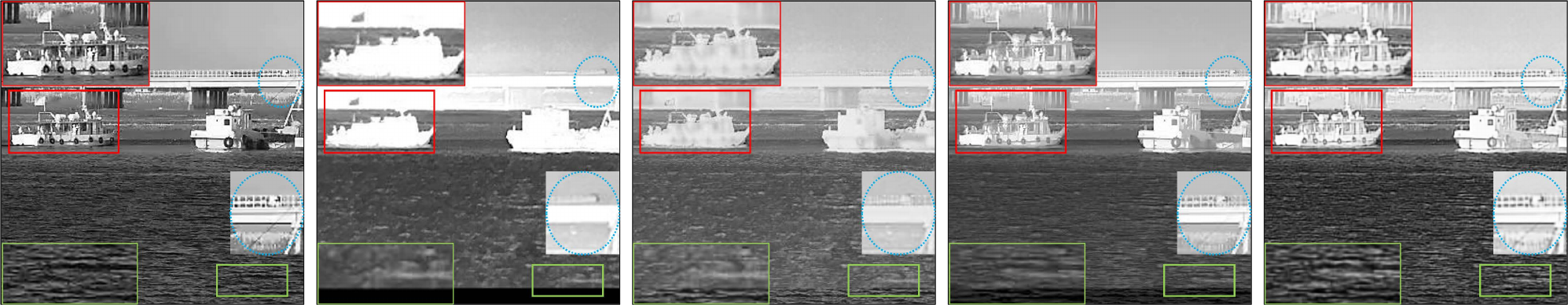}\\
    \begin{flushleft}  {\fontsize{8pt}{10pt}\selectfont  
    \vspace{-5pt}
    \hspace{0.43cm} SWIR \hspace{0.9cm} LWIR \hspace{0.75cm} DATFuse \hspace{0.58cm} DDFM \hspace{0.75cm} EMMA \\} \end{flushleft} 
    \vspace{-1pt}
    \includegraphics[width=0.47\textwidth]{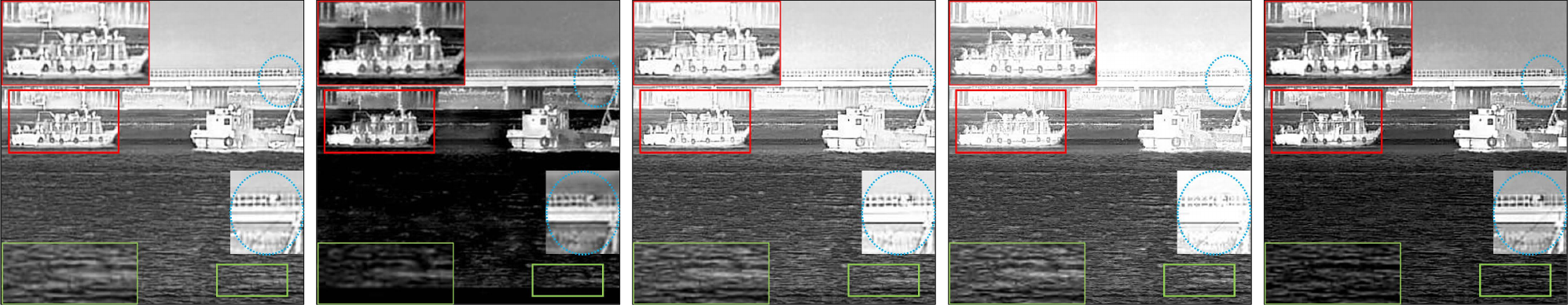}\\
    \begin{flushleft}  {\fontsize{8pt}{10pt}\selectfont  
    \vspace{-5pt}
    \hspace{0.25cm} DifFusion \hspace{0.65cm} IGNet \hspace{0.55cm} SeAFusion \hspace{0.3cm} SwinFusion \hspace{0.55cm} ours \\} \end{flushleft} 
    \vspace{-1pt}
    \caption{Qualitative comparisons of various methods on several images from the NSLSR dataset.}\label{fig:NS}
    \Description{NSLSR_1}
\end{figure}
\begin{figure}[htbp]
    \includegraphics[width=0.47\textwidth]{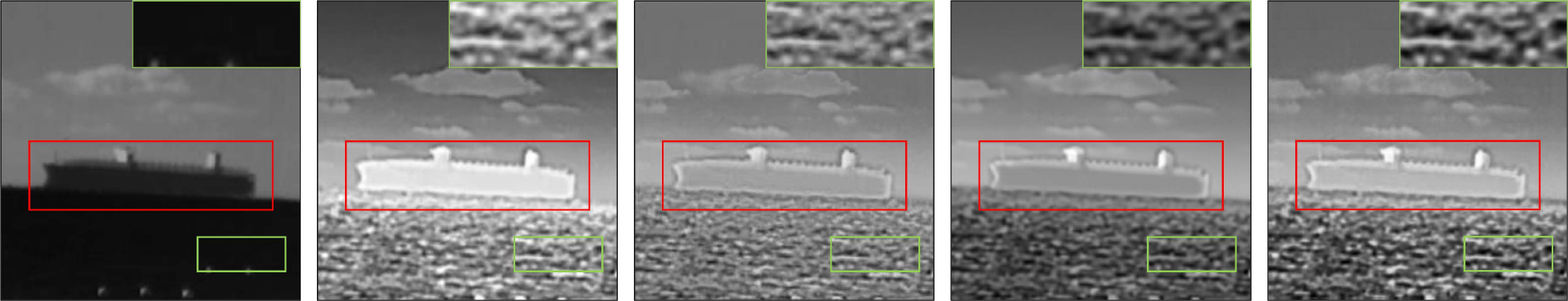}\\
    \begin{flushleft}  {\fontsize{8pt}{10pt}\selectfont  
    \vspace{-5pt}
    \hspace{0.43cm} SWIR \hspace{0.9cm} LWIR \hspace{0.75cm} DATFuse \hspace{0.58cm} DDFM \hspace{0.75cm} EMMA \\} \end{flushleft}
    \vspace{-1pt}
    \includegraphics[width=0.47\textwidth]{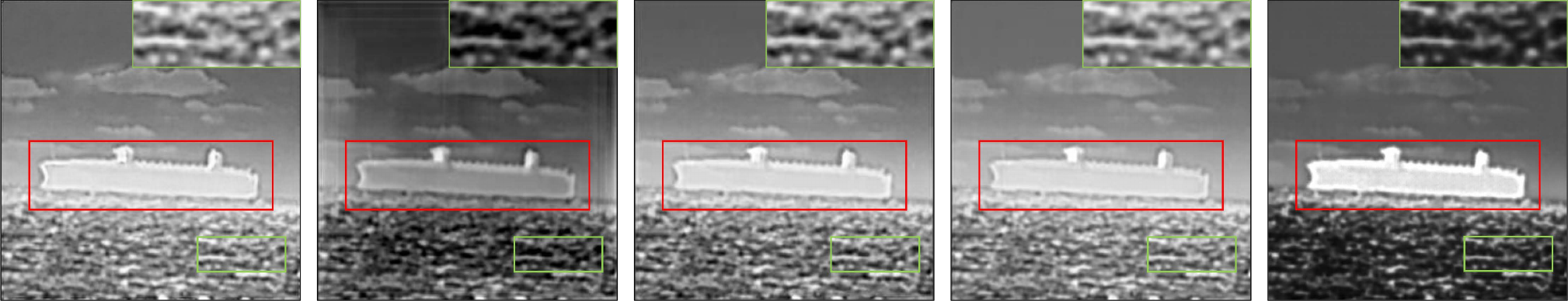}\\
    \begin{flushleft}  {\fontsize{8pt}{10pt}\selectfont  
    \vspace{-5pt}
    \hspace{0.25cm} DifFusion \hspace{0.65cm} IGNet \hspace{0.55cm} SeAFusion \hspace{0.3cm} SwinFusion \hspace{0.55cm} ours \\} \end{flushleft} 
    \vspace{-1pt}
    \includegraphics[width=0.47\textwidth]{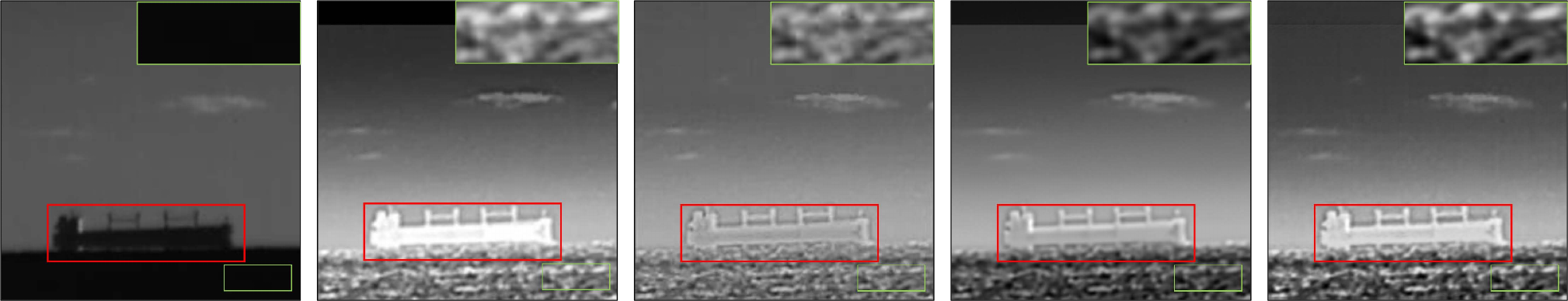}\\
    \begin{flushleft}  {\fontsize{8pt}{10pt}\selectfont  
    \vspace{-5pt}
    \hspace{0.43cm} SWIR \hspace{0.9cm} LWIR \hspace{0.75cm} DATFuse \hspace{0.58cm} DDFM \hspace{0.75cm} EMMA \\} \end{flushleft} 
    \vspace{-1pt}
    \includegraphics[width=0.47\textwidth]{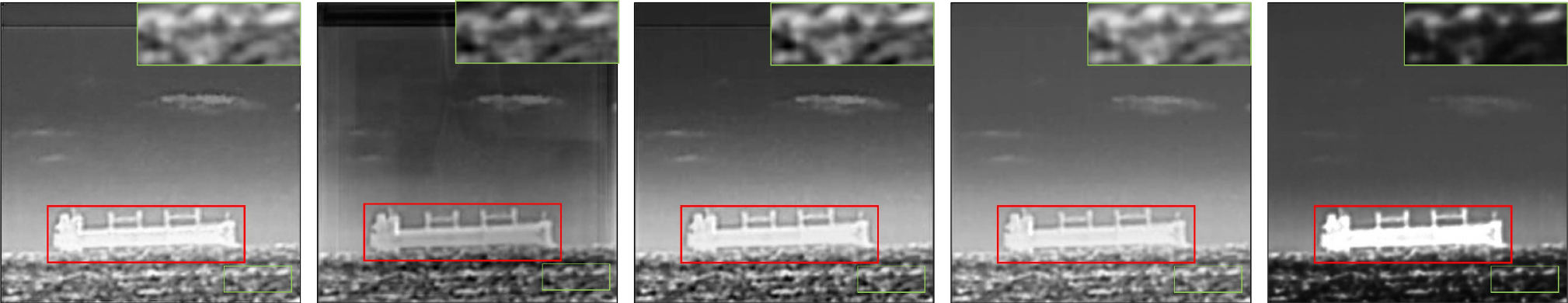}\\
    \begin{flushleft}  {\fontsize{8pt}{10pt}\selectfont  
    \vspace{-5pt}
    \hspace{0.25cm} DifFusion \hspace{0.65cm} IGNet \hspace{0.55cm} SeAFusion \hspace{0.3cm} SwinFusion \hspace{0.55cm} ours \\} \end{flushleft} 
    \vspace{-1pt}
    \caption{Qualitative comparison of various methods on several images from the IRD dataset.}\label{fig:IRD}
    \Description{NSLSR_2}
\end{figure}
\begin{table*}[t]
\centering  
\caption{Quantitative comparisons of SOTA fusion methods on NSLSR and ISD datasets. We mark the best result in deep red, the second-best result in deep blue, and the third-best result will be \underline{underlined}.}  
\label{tab:nslsr_results}  
\begin{tabular}{l|cccccc|cccccc}  
\toprule
 \multicolumn{1}{c|}{\multirow{2}{*}{\textbf{Methods}}} & \multicolumn{6}{c|}{\textbf{NSLSR}} & \multicolumn{6}{c}{\textbf{ISD}} \\
                  & \textbf{EN$\uparrow$} & \textbf{SF$\uparrow$} & \textbf{SD$\uparrow$} & \textbf{SCD$\uparrow$} & \textbf{VI$\uparrow$} & \textbf{Qabf$\uparrow$} & \textbf{EN$\uparrow$} & \textbf{SF$\uparrow$} & \textbf{SD$\uparrow$} & \textbf{SCD$\uparrow$} & \textbf{VI$\uparrow$} & \textbf{Qabf$\uparrow$} \\ \hline
DATFuse (TCSVT2023) \cite{Tang_2023_DATFuse} & 6.845 & 16.498   & 44.814 & 0.297 & \underline{0.567} & 0.431 & 6.474 & 10.115 & 28.251 & 1.254 & 0.587 & 0.579 \\
EMMA (cvpr2024) \cite{Zhao_2024_CVPR} & \underline{7.122} & 19.572  & \underline{62.776} & \textcolor{darkblue}{\textbf{1.405}} & \textcolor{darkblue}{\textbf{0.589}} & \underline{0.501} & \underline{7.167} & 12.130 & \underline{43.894} & \underline{1.675} & \underline{0.836} & 0.616 \\
DDFM (ICCV2023) \cite{Zhao_2023_ICCV} & 6.989 & 14.278  & 54.587 & 1.189 & 0.538 & 0.394 & 7.103 & 9.041 & 42.720 & \textcolor{darkred}{\textbf{1.836}} & 0.742 & 0.555 \\
DifFusion (TIP2023) \cite{Diffusioncolor} & \textcolor{darkred}{\textbf{7.216}} & \underline{20.522}  & 56.576 & 0.732 & 0.542 & \textcolor{darkblue}{\textbf{0.514}} & 7.111 & \textcolor{darkred}{\textbf{12.914}} & 43.830 & 1.154 & 0.740 & 0.646 \\
SeAFusion (IF2022) \cite{TANG202228} & 7.001 & \textcolor{darkblue}{\textbf{20.678}}  & \textcolor{darkblue}{\textbf{62.936}} & 1.165 & 0.557 & 0.480 & \textcolor{darkred}{\textbf{7.306}} & \textcolor{darkblue}{\textbf{12.854}} & \textcolor{darkblue}{\textbf{49.407}} & 1.509 & 0.829 & \textcolor{darkblue}{\textbf{0.655}} \\
SwinFusion (JAS2022) \cite{9812535} & 6.654 & 18.978  & 61.699 & \underline{1.204} & 0.552 & 0.462 & 7.003 & 11.180 & 42.702 & 1.331 & \textcolor{darkblue}{\textbf{0.841}} & \textcolor{darkred}{\textbf{0.662}} \\
IGNet (ACMMM2023) \cite{li2023learning} & 6.683 & 16.848  & 57.874 & 1.064 & 0.535 & 0.399 & 7.090 & 11.853 & 40.984 & 1.210 & 0.632 & 0.590 \\
LSFDNet (ours) & \textcolor{darkblue}{\textbf{7.181}} & \textcolor{darkred}{\textbf{21.022}}  & \textcolor{darkred}{\textbf{64.723}} & \textcolor{darkred}{\textbf{1.427}} & \textcolor{darkred}{\textbf{0.611}} & \textcolor{darkred}{\textbf{0.520}} & \textcolor{darkblue}{\textbf{7.173}} & \underline{12.330} & \textcolor{darkred}{\textbf{50.340}} & \textcolor{darkblue}{\textbf{1.687}} & \textcolor{darkred}{\textbf{0.848}} & \underline{0.651} \\
\bottomrule
\end{tabular}  
\end{table*} 
\subsection{Results of SWIR and LWIR Image Fusion} \label{sec:fusion}
Since there are no multimodal image fusion algorithms specifically designed for SWIR and LWIR images, we select six state-of-the-art (SOTA) general multimodal fusion algorithms or visible-infrared image fusion algorithms developed in recent years, including DATFuse \cite{Tang_2023_DATFuse}, DDFM \cite{Zhao_2024_CVPR}, EMMA \cite{Zhao_2023_ICCV}, DifFusion \cite{Diffusioncolor}, IGNet \cite{TANG202228}, SeAFusion \cite{9812535} and  SwinFusion \cite{li2023learning}. We evaluate the performance of our LSFDNet by comparing it with these algorithms.

\textbf{Qualitative results.} For ship fusion detection, our goal is to ensure that the fused image contains rich ship information with high contrast. The results of different methods on the NSLSR dataset are shown in Fig. \ref{fig:NS}. Compared to other methods, our proposed method demonstrates two significant advantages. First, as observed in the green boxes, our method is notably more effective at suppressing sea surface noise compared to other algorithms. This results in higher image contrast and makes the contours of the ship more distinct. Second, as shown in the red boxes and blue circles, our method better preserves the information of ship, maintaining finer details and a more reasonable brightness distribution. From the figure, it is also clear that IGNet performs well in reducing sea surface noise, but as indicated in the red boxes, IGNet also reduces the retention of object information. Moreover, while DDFM, which uses a diffusion model \cite{ho2020denoising}, achieves lower overall noise in the image, some detailed information of the ship (as shown in the blue circle) is removed. The qualitative results of these methods on the IRD dataset are shown in Fig. \ref{fig:IRD}, which further corroborates the two distinct advantages of our algorithm. From these results, it can be concluded that LSFDNet effectively suppresses sea surface noise while retaining object information to the greatest extent, resulting in superior visual quality.

\begin{figure}[t]
    \includegraphics[width=0.47\textwidth]{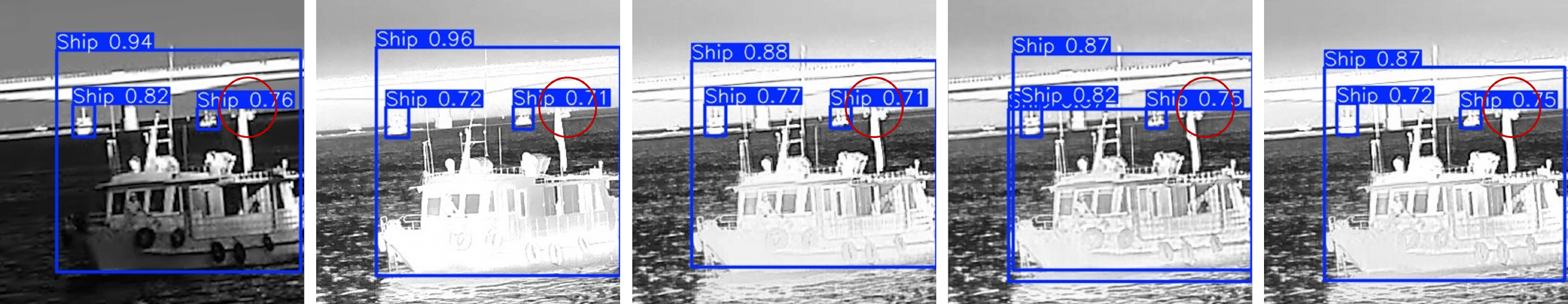}\\
    \begin{flushleft}  {\fontsize{8pt}{10pt}\selectfont  
    \vspace{-5pt}
    \hspace{0.5cm} SWIR \hspace{0.88cm} LWIR \hspace{0.58cm} SeAFusion \hspace{0.4cm} DifFusion \hspace{0.31cm} SwinFusion \\} \end{flushleft}
    \vspace{-1pt}
    \includegraphics[width=0.47\textwidth]{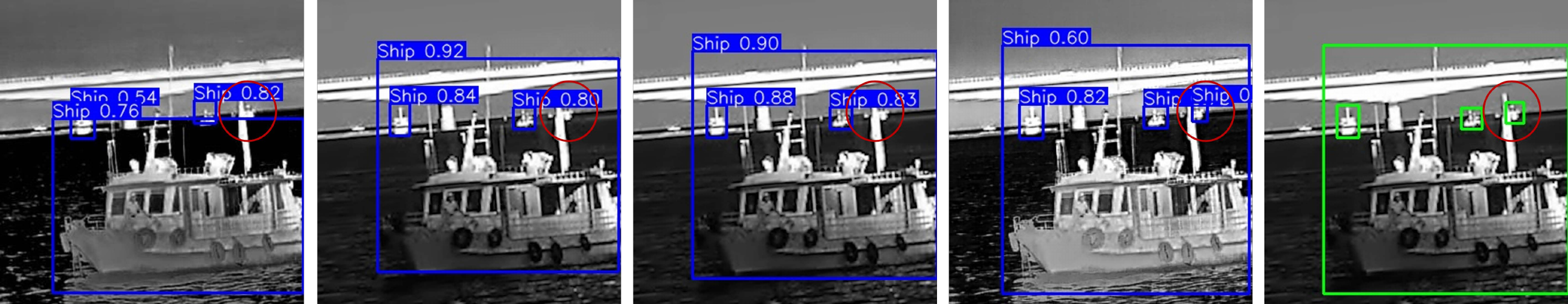}\\
    \begin{flushleft}  {\fontsize{8pt}{10pt}\selectfont  
    \vspace{-5pt}
    \hspace{0.45cm} IGNet \hspace{0.5cm} CAFF-DINO \hspace{0.33cm} DEYOLO \hspace{0.5cm} LSFDNet \hspace{0.80cm} GT \\} \end{flushleft} 
    \vspace{-1pt}
    \caption{Qualitative comparisons of the detection performance on the NSLSR dataset.}\label{fig:Det}
    \Description{NSLSR_2}
\end{figure}

\begin{table}[t]
\centering  
\caption{Quantitative comparisons of the detection performance on NSLSR datasets. SW/LW and MF$+$OD tasks utilize YOLOv12 as the detection network. We mark the best result in deep red, the second-best result in deep blue.}  
\label{tab:results}  
\begin{tabular}{l|l|cccc}  
\toprule  
\textbf{Task} & \textbf{Methods} & \textbf{P} & \textbf{R} & \textbf{mAP$_{50}$} & \textbf{mAP$_{50:95}$} \\
\hline  
\multirow{2}{*}{SW/LW}   
  & SWIR & 0.911 & 0.893 & 0.942 & 0.666 \\
  & LWIR & 0.903 & 0.870 & 0.929 & 0.628 \\
\hline
\multirow{4}{*}{MF+OD}   
  & SeA      & 0.920 & 0.864 & 0.943 & 0.689 \\
  & IGNet    & 0.946 & 0.851 & 0.942 & 0.676 \\
  & Fusiondif& 0.925 & 0.889 & 0.952 & 0.698 \\
  & Swin     & 0.918 & 0.881 & 0.949 & 0.674 \\
\hline 
\multirow{2}{*}{MOD}   
  & CAFF-DINO & 0.925 & 0.905 & \textcolor{darkblue}{\textbf{0.958}} & \textcolor{darkblue}{\textbf{0.706}} \\
  & DEYOLO & 0.938 & 0.885 & 0.956 & 0.702 \\
\hline 
MF-OD & LSFDNet & 0.934 & 0.887 & \textcolor{darkred}{\textbf{0.962}} & \textcolor{darkred}{\textbf{0.770}} \\
\bottomrule
\end{tabular}
\end{table}  
\begin{table*}
\centering  
\caption{Quantitative ablation experiment results of the proposed OE loss and MLCF Module. The best results are highlighted in deep red, while the second-best results are highlighted in deep blue.}  
\begin{tabular}{c|c|ccc|ccccccc}  
\toprule  
\multirow{2}{*}{\textbf{Model}} & \multirow{2}{*}{\textbf{OE Loss}} & \multicolumn{3}{c|}{\textbf{MLCF}} & \multicolumn{6}{c}{\textbf{NSLSR}} \\
& & \textbf{Multimodal} & \textbf{Multiscale} & \textbf{Multiask} & \textbf{EN$\uparrow$} & \textbf{SF$\uparrow$} & \textbf{SD$\uparrow$} & \textbf{SCD$\uparrow$} & \textbf{VI$\uparrow$} & \textbf{Qabf$\uparrow$} \\
\hline 
M1 & $\times$ & $\times$ & $\times$ & $\times$ & 6.687 & 18.365 & 58.268 & 1.023 & 0.566 & 0.431 \\
M2 & $\times$ & $\checkmark$ & $\checkmark$ & $\checkmark$ & \textcolor{darkblue}{\textbf{7.109}} & \textcolor{darkblue}{\textbf{21.012}} & 62.392 & \textcolor{darkblue}{\textbf{1.387}} & 0.603 & 0.534 \\
M3 & $\checkmark$ & $\times$ & $\times$ & $\times$ & 6.785 & 19.675 & 60.053 & 1.145 & 0.557 & 0.490 \\
M4 & $\checkmark$ & $\times$ & $\checkmark$ & $\checkmark$ & 6.915 & 19.863 & \textcolor{darkblue}{\textbf{64.321}} & 1.219 & 0.563 & 0.479 \\
M5 & $\checkmark$ & $\checkmark$ & $\times$ & $\checkmark$ & 7.043 & 20.169 & 62.806 & 1.371 & 0.583 & 0.516 \\
M6 & $\checkmark$ & $\checkmark$ & $\checkmark$ & $\times$ & 7.049 & 20.893 & 63.758 & 1.363 & \textcolor{darkred}{\textbf{0.614}} & \textcolor{darkred}{\textbf{0.507}} \\
M7(LSFDNet) & $\checkmark$ & $\checkmark$ & $\checkmark$ & $\checkmark$ & \textcolor{darkred}{\textbf{7.181}} & \textcolor{darkred}{\textbf{21.022}} & \textcolor{darkred}{\textbf{64.723}} & \textcolor{darkred}{\textbf{1.427}} & \textcolor{darkblue}{\textbf{0.611}} & \textcolor{darkblue}{\textbf{0.520}} \\
\bottomrule  
\end{tabular}  
\label{table:model_performance}  
\end{table*} 

\begin{table}[t]
\centering  
\caption{Quantitative ablation experiment results of the $F_{f}$ in Detection. The best results are highlighted.}  
\label{tab:ablation det}  
\begin{tabular}{c|cccccccc}  
\toprule  
\textbf{Methods} & \textbf{P} & \textbf{R} & \textbf{mAP$_{50}$} & \textbf{mAP$_{50:95}$} \\
\hline  
w/o $F_{f}$ & 0.905 & 0.854 & 0.953 & 0.725 \\
LSFDNet(ours)  & 0.934 & 0.887 & \textbf{0.962} & \textbf{0.770} \\
\bottomrule
\end{tabular}
\end{table}  
\textbf{Quantitative results.} As shown in Table \ref{tab:nslsr_results}, we compare LSFDNet with six state-of-the-art algorithms across two datasets, and our algorithm achieves the highest or second-highest average scores across six metrics. Notably, the outstanding scores for $SD$ and $VIF$ demonstrate that our method maintains high contrast and excellent visual fidelity. Even after suppressing sea surface noise, LSFDNet still achieves high scores for $SF$ and $Qabf$, demonstrating our method’s ability to preserve texture information effectively. Similarly, the outstanding $SCD$ and $EN$ metrics indicate that our results possess a high level of information richness. On the NSLSR dataset, our method performs particularly well, effectively retaining complex texture details and adapting to intricate coastal backgrounds. However, on the ISD dataset, the performance is somewhat less impressive, primarily due to the overly monotonous nature of the images. The partial suppression of sea surface noise in this dataset leads to significant information loss, impacting the overall results.
\subsection{Results of Multimodal Object Detection} \label{sec:detection}
To comprehensively evaluate the detection performance of the proposed LSFDNet, we conducted a comparison using multiple approaches, including single-modal LW/SW image detection, several fused image detection, and specialized multimodal object detection (MOD) algorithms \cite{HelvigCAFFDINO2024, Chen_2024_ICPR}. Since the detection component of the LSFDNet, designed for single-stage multimodal image fusion and object detection (MF$-$OD) task, is based on YOLOv12s framework, we utilize YOLOv12s as the detection network for SWIR/LWIR image fusion and MF$+$OD tasks.

\textbf{Qualitative results.} The detection results of different methods on the NSLSR dataset are visualized in Fig. \ref{fig:Det}. Our method achieves visually superior detection results. In the ground truth (GT) image, the red box highlights a small ship in the distance, which is partially obscured by the mast of a larger, nearby ship. This represents a weak and incomplete object, making its detection particularly challenging. Among the methods tested, only our approach successfully detects the distant boat. This success is attributed to the network design, which effectively integrates multimodal image information and significantly reduces missed detections.

\textbf{Quantitative Results.} Table \ref{tab:results} presents the performance of various detection methods on the NSLSR dataset. It can be observed that methods leveraging fused images or multi-modal information outperform single-modal object detection. This is because single-modal images carry less information compared to multimodal images, especially when certain single-modal inputs suffer from severe information loss. Furthermore, methods specifically designed for multimodal detection perform better than those that detect objects from fused images, as these approaches can comprehensively extract features. LSFDNet, in particular, aggregates fused features in addition to leveraging multimodal characteristics, further enriching the feature set for detection. These additional features enables our algorithm to achieve the best performance among all methods. Specifically, our approach achieves a significant improvement in mAP$_{50:95}$, surpassing other methods by \textbf{$10\%$}.

\subsection{Ablation studies} \label{sec:Ablation}
\textbf{Effects of the OE Loss and Multi-Level Cross-Fusion (MLCF) Module.} The MLCF Module mainly consists of multimodal, multiscale and multitask feature attention blocks. To evaluate the effectiveness of the OE loss and the MLCF module, these components are progressively removed from the network. Table \ref{table:model_performance} details the ablation experiments conducted on the NSLSR dataset. A comparison between M$2$ and M$7$ shows a noticeable drop in the $SD$ metric after removing the OE loss, with other metrics experiencing slight decreases. This indicates that the OE loss encourage the network to focus on ships, thereby improving the visual quality of the fused images. The comparison between M$3$ and M$7$ underscores the critical role of the MLCF module, as its absence leads to significant metric declines, demonstrating its ability to effectively fusion more features. Additionally, comparisons of M$4$, M$5$, M$6$ and M$7$ confirm the positive impact of each feature aggregation block in the MLCF module. Notably, multimodal feature aggregation is crucial, while simple concatenation and decoding yield suboptimal results. Moreover, the experimental results from M6 indicate that the detection features from the detection network can enhance the performance of the fusion network.

\textbf{Effect of the Fusion Feature $F_{f}$.} The detection network utilizes the aggregated fusion features $F_{f}$ from the fusion task. We remove $F_{f}$ from the shallow feature extraction stage of the detection network and directly aggregate the detection features $F_{LS}$ and $F_{WS}$. As shown in Table \ref{tab:ablation det}, the results of the ablation experiment indicate that the removal of the fusion features leads to a decline in both mAP$_{50}$ and mAP$_{50:95}$. This demonstrates that the fusion features effectively enrich detection semantics and enhance detection performance.

\section{Conclusion}
\label{sec:conclusion}
This paper presents LSFDNet, an innovative approach for robust maritime ship detection through SWIR-LWIR image fusion. By leveraging an end-to-end network architecture, LSFDNet integrates feature extraction, fusion, and detection tasks, enhancing both visual quality and detection accuracy. The proposed Multi-Level Cross-Fusion (MLCF) module seamlessly combines multimodal, multiscale, and multitask features, while task-specific loss functions, like Object Enhancement (OE) loss, further improve target focus. We also introduce the Nearshore Ship Long-Short Wave Registration Dataset (NSLSR), tailored for SWIR-LWIR maritime ship detection, advancing research in this domain. Overall, LSFDNet provides a promising solution for ship detection in complex maritime environments, offering both superior fusion performance and highly accurate detection capabilities. 
\begin{acks}
This research is supported by the Ningbo 2025 Science \& Technology Innovation Major Project (No. 2023Z044), and the Singapore Ministry of Education (MOE) Academic Research Fund (AcRF) Tier 1 grant (22-SIS-SMU-093).
\end{acks}

\bibliographystyle{ACM-Reference-Format}
\bibliography{sample-base}




\end{document}